\documentclass{sig-alternate}

\usepackage{graphicx}   
\usepackage{subfigure} 
\usepackage{color}
\usepackage{url}

\begin{document}
%

\CopyrightYear{2016} 
\setcopyright{rightsretained} 
\conferenceinfo{MM '16}{October 15-19, 2016, Amsterdam, Netherlands} 
\isbn{978-1-4503-3603-1/16/10}
\doi{http://dx.doi.org/10.1145/2964284.2973791}

\title{LightNet: A Versatile, Standalone Matlab-based Environment for Deep Learning}
\subtitle{[Simplify Deep Learning in Hundreds of Lines of Code]}
\author{Chengxi Ye, Chen Zhao*, Yezhou Yang, Cornelia Ferm\"{u}ller, Yiannis Aloimonos
\\
Computer Science Department, University of Maryland, College Park, MD 20740, USA. 
\\
\{cxy, yzyang, fer, yiannis\}@umiacs.umd.edu *chenzhao@umd.edu}



\maketitle
\begin{abstract}
LightNet is a \textbf{lightweight}, \textbf{versatile}, \textbf{purely Matlab-based} deep learning framework. The idea underlying its design is to provide an easy-to-understand, easy-to-use and efficient computational platform for deep learning research. The implemented framework supports major deep learning architectures such as Multilayer Perceptron Networks (MLP), Convolutional Neural Networks (CNN) and Recurrent Neural Networks (RNN). The framework also supports both CPU and GPU computation, and the switch between them is straightforward. Different applications in computer vision, natural language processing and robotics are demonstrated as experiments.

\textbf{Availability}: the source code and data is available at: \url{https://github.com/yechengxi/LightNet}

\end{abstract}




\section{Introduction}

Deep neural networks ~\cite{krizhevsky2012imagenet} have given rise to major advancements in many problems of machine intelligence. Most current implementations of neural network models primarily emphasize efficiency. These pipelines (Table ~\ref{tab:Packages}) can consist of a quarter to half a million lines of code and often involve multiple programming languages ~\cite{jia2014caffe,vedaldi2015matconvnet,bastien2012theano}. It requires extensive efforts to thoroughly understand and modify the models. A straightforward and self-explanatory deep learning framework is highly anticipated to accelerate the understanding and application of deep neural network models.

\begin{table}[]
\scriptsize
\centering
\caption{Deep Neural Network Packages}
\label{tab:Packages}
\begin{tabular}{|l|r|r|r|}
\hline
\multicolumn{1}{|c|}{Framework} & \multicolumn{1}{c|}{Language} & \multicolumn{1}{c|}{Native Models} & \multicolumn{1}{c|}{Lines of Code} \\ \hline
Caffe                           & C++                           & CNN                                & 74,903                             \\ \hline
Theano                          & Python, C                        & MLP/CNN/RNN                        & 148,817                            \\ \hline
Torch                           & Lua, C                           & MLP/CNN/RNN                        & 458,650                            \\ \hline
TensorFlow                      & C++                           & MLP/CNN/RNN                        & 335,669                            \\ \hline
Matconvnet                      & Matlab, C                     & CNN                                & 43,087                             \\ \hline
LightNet                        & Matlab                        & MLP/CNN/RNN                        & 951 (1,762)*                        \\ \hline

\end{tabular}
\parbox{\linewidth}{\scriptsize%
\textsc{*}
Lines of code in the core modules and in the whole package.}      
\end{table}

We present LightNet, a \textbf{lightweight}, \textbf{versatile}, \textbf{purely Matlab-based} implementation of modern deep neural network models. Succinct and efficient Matlab programming techniques have been used to implement all the computational modules. Many popular types of neural networks, such as multilayer perceptrons, convolutional neural networks, and recurrent neural networks are implemented in LightNet, together with several variations of stochastic gradient descent (SDG) based optimization algorithms. 

Since LightNet is implemented \textbf{solely} with Matlab, the major computations are vectorized and implemented in \textbf{hundreds} of lines of code, orders of magnitude more succinct than existing pipelines. All fundamental operations can be easily customized, only basic knowledge of Matlab programming is required. Mathematically oriented researchers can focus on the mathematical modeling part rather than the engineering part. Application oriented users can easily understand and modify any part of the framework to develop new network architectures and adapt them to new applications. Aside from its simplicity, LightNet has the following features: 
1. LightNet contains the most modern network architectures.
2. Applications in computer vision, natural language processing and reinforcement learning are demonstrated.
3. LightNet provides a comprehensive collection of optimization algorithms.
4. LightNet supports straightforward switching between CPU and GPU computing.
5. Fast Fourier transforms are used to efficiently compute convolutions, and thus large convolution kernels are supported.
6. LightNet automates hyper-parameter tuning with a novel Selective-SGD algorithm. 

\section{Using the Package}

\begin{figure}
\centering
\includegraphics[height=2.9in]{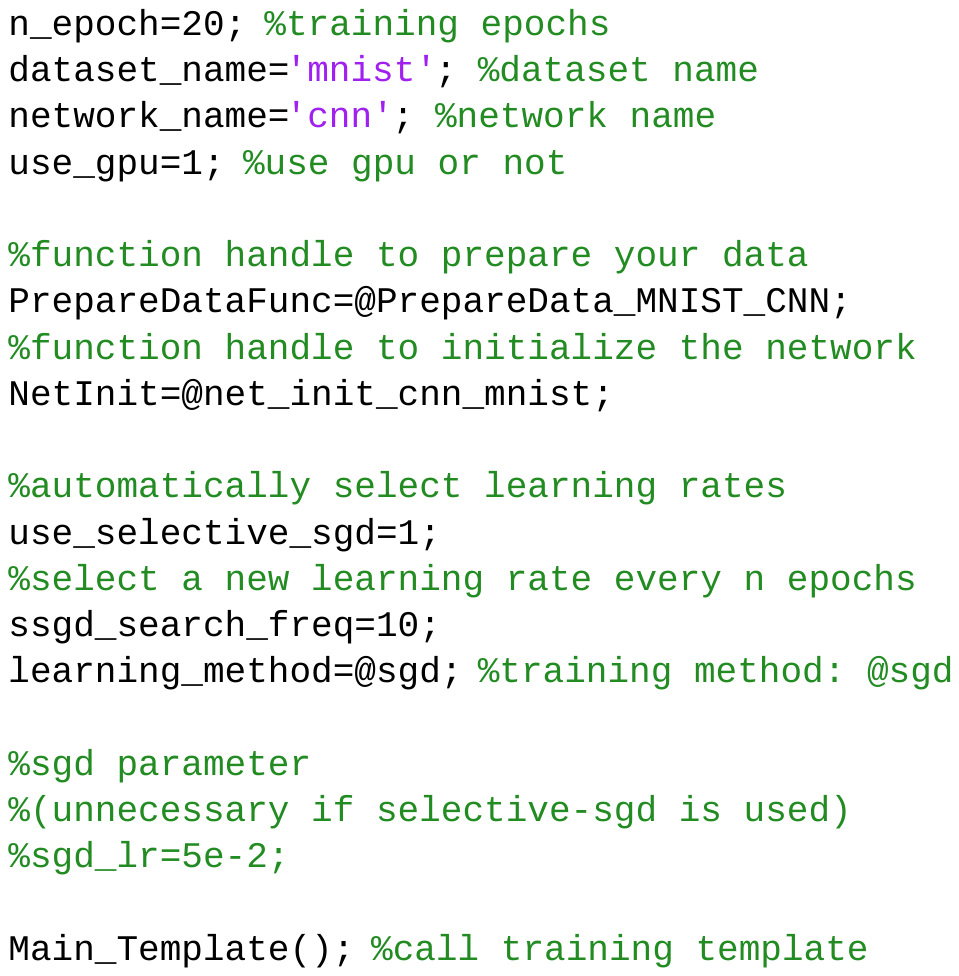}
\caption{A basic example, which shows how to train a CNN on the MNIST dataset with LightNet.}
\label{fig:RunningTemplate}
\end{figure}

An example of using LightNet can be found in (Fig. ~\ref{fig:RunningTemplate}): a simple template is provided to start the training process. The user is required to fill in some critical training parameters, such as the number of training epochs, or the training method. A Selective-SGD algorithm is provided to facilitate the selection of an optimal learning rate. The learning rate is selected automatically, and can optionally be adjusted during the training.
The framework supports both GPU and CPU computation, through the $opts.use\_gpu$ option. Two additional functions are provided to prepare the training data and initialize the network structure. Every experiment in this paper can reproduced by running the related script file. More details can be found on the project webpage.

\section{Building Blocks}

The primary computational module includes a feed forward process and a backward/back propagation process.
The feed forward process evaluates the model, and the back propagation reports the network gradients. Stochastic gradient descent based algorithms are used to optimize the model parameters.

\subsection{Core Computational Modules}

LightNet allows us to focus on the mathematical modeling of the network, rather than low-level engineering details. To make this paper self-contained, we explain the main computational modules of  LightNet. All networks ( and related experiments) in this paper are built with these modules. The notations below are chosen for simplicity. Readers can easily extend the derivations to the mini-batch setting.

\subsubsection{Linear Perceptron Layer}
\label{sec:linear}

A linear perceptron layer can be expressed as: $y= W x + b$. Here, $x$ denotes the input data of size $input\_dim \times 1$, $W$ denotes the weight matrix of size $output\_dim \times input\_dim$, $b$ is a bias vector of size $output\_dim \times 1$, and $y$ denotes the linear layer output of size $output\_dim \times 1$. 

The mapping from the input of the linear perceptron to the final network output can be expressed as: $z=f(y)=f(W x + b)$, where $f$ is a non-linear function that represents the network's computation in the deeper layers, and $z$ is the network output, which is usually a loss value.

The backward process calculates the derivative $\frac{\partial z}{\partial x}$, which is the derivative passing to the shallower layers, and $\frac{\partial z}{\partial W}$, $\frac{\partial z}{\partial b}$, which are the gradients that guide the gradient descent process.
\begin{equation}
\frac{\partial z}{\partial x} = \frac{\partial z}{\partial y} \cdot \frac{\partial y}{\partial x} = f'(y)^{T} \cdot W
\label{dzdx}
\end{equation}
\begin{equation}
\frac{\partial z}{\partial W} = \frac{\partial z}{\partial y} \cdot \frac{\partial y}{\partial W} = f'(y) \cdot x^{T}
\label{dzdw}
\end{equation}
\begin{equation}
\frac{\partial z}{\partial b} = \frac{\partial z}{\partial y} \cdot \frac{\partial y}{\partial b}  = f'(y)
\label{dzdb}
\end{equation}

The module adopts extensively optimized Matlab matrix operations to calculate the matrix-vector products.

\subsubsection{Convolutional Layer}

A convolutional layer maps $N_{map\_in}$ input feature maps to $N_{map\_out}$ output feature maps with a multidimensional filter bank $k_{io}$. Each input feature map $x_{i}$ is convolved with the corresponding filter bank $k_{io}$. The convolution results are summed, and  a bias value $b_{o}$ is added, to generate the $o$-th output map: $y_o=\sum_{1 \leq i \leq N_{map\_in} }{ k_{io} * x_i  }  + b_o$. To allow using large convolution kernels, fast Fourier transforms (FFT) are used for computing convolutions (and correlations). According to the convolution theorem ~\cite{mallat2008wavelet}, convolution in the spatial domain is equivalent to point-wise multiplication in the frequency domain. Therefore, $k_i *x_i$ can be calculated using the Fourier transform as: $k_i *x_i=\mathcal{F}^{-1} \{  \mathcal{F}\{ k_i \} \cdot  \mathcal{F}\{ x_i \}  \} $. Here, $\mathcal{F}$ denotes the Fourier transform and $\cdot$ denotes the point-wise multiplication operation. The convolution layer supports both padding and striding. 

The mapping from the $o$-th output feature map to the network output can be expressed as: $z=f(y_o)$. Here $f$ is the non-linear mapping from the $o$-th output feature map $y_o$ to the final network output. As before (in Sec.~\ref{sec:linear}), $\frac{\partial z}{\partial x_i}$, $\frac{\partial z}{\partial k_i}$, and $\frac{\partial z}{\partial b_o}$ need to be calculated in the backward process, as follows:
\begin{equation}
\frac{\partial z}{\partial x_i} = \frac{\partial z}{\partial y_o} \cdot \frac{\partial y_o}{\partial x_i} = f'(y_o
) \star k_i,
\label{dzdxi}
\end{equation}

\noindent where $\star$ denotes the correlation operation. Denoting the complex conjugate as  $conj$, this correlation is calculated  in the frequency domain using the Fourier transform as: 
$ x \star k = \mathcal{F}^{-1} \{ \mathcal{F}\{ x \} \cdot conj ( \mathcal{F}\{ k \} ) \}$.
\begin{equation}
\frac{\partial z}{\partial k_{io}^{*}} = \frac{\partial z}{\partial y_o} \cdot \frac{\partial y_o}{\partial k_{io}^{*}} = f'(y_o
) \star x_i,
\label{dzdkio}
\end{equation}

\noindent where $k^{*}$ represents the flipped  kernel $k$. Thus, the gradient $\frac{\partial z}{\partial k_{io}}$ is  calculated by flipping the correlation output. Finally,
\begin{equation}
\frac{\partial z}{\partial b_o} = \frac{\partial z}{\partial y_o} \cdot \frac{\partial y_o}{\partial b_o} = 1^{T}\cdot vec(f'(y_o)) 
\label{dzdbo}
\end{equation}

\noindent In words, the gradient $\frac{\partial z}{\partial b_o}$ can be calculated by point-wise summation of the values in $f'(y_o)$. 

\subsubsection{Max-pooling Layer}

The max pooling layer calculates the largest element in $P_{r} \times P_{c}$ windows, with stride size $S_{r} \times S_{c}$.  A customized $im2col\_ln$ function is implemented to convert the stridden pooling patches into column vectors, to vectorize the pooling computation in Matlab. The built-in $max$ function is called on these column vectors to return the pooling result and the indices of these maximum values. Then, the indices in the original batched data are recovered accordingly. Also, zero padding can be applied to the input data.

Without the loss of generality, the mapping from the max-pooling layer input to the final network output can be expressed as: $z=f(y)=f( S x )$, where $S$ is a selection matrix, and $x$ is a column vector which denotes the input data in this layer. 

In the backward process, $\frac{\partial z}{\partial x}$ is calculated and passed to the shallower layers: $\frac{\partial z}{\partial x} = \frac{\partial z}{\partial y} \cdot S = f'(y)^{T} S$.

When the pooling range is less than or equal to the stride size, $\frac{\partial z}{\partial x}$ can be calculated with simple matrix indexing techniques in Matlab. Specifically, an empty tensor $dzdx$ of the same size with the input data is created. $dzdx(from)=dzdy$, where $from$ is the pooling indices, and $dzdy$ is a tensor recording the pooling results. When the pooling range is larger than the stride size, each entry in $x$ can be pooled multiple times, and the back propagation gradients need to be accumulated for each of these multiple-pooled entries. In this case, the $\frac{\partial z}{\partial x}$ is calculated using the Matlab function: $accumarray()$.

\subsubsection{Rectified Linear Unit}

The rectified linear unit ($ReLU$) is implemented as a major non-linear mapping function, some other functions including $sigmoid$ and $tanh$ are omitted from the discussion here. The $ReLU$ function is the identity function if the input is larger than $0$ and outputs $0$ otherwise: $y = relu(x)=x \cdot ind(x>0)$. In the backward process, the gradient is passed to the shallower layer if the input data is non-negative. Otherwise, the gradient is ignored.

\subsection{Loss function}

Usually, a loss function is connected to the outputs of the deepest core computation module. Currently, LightNet supports the softmax log-loss function for classification tasks.

\subsection{Optimization Algorithms}

Stochastic gradient descent (SGD) algorithm based optimization algorithms are the primary tools to train deep neural networks. The standard SGD algorithm and several of its popular variants such as Adagrad~\cite{duchi2011adaptive}, RMSProp~\cite{tieleman2012lecture} and Adam~\cite{kingma2014adam} are also implemented for deep learning research. It is worth mentioning that we implement a novel Selective-SGD algorithm to facilitate the selection of hyperparameters, especially the learning rate. This algorithm selects the most efficient learning rate by running the SGD process for a few iterations using each learning rate from a discrete candidate set. During the middle of the neural net training, the Selective-SGD algorithm can also be applied to select different learning rates to accelerate the energy decay.

\section{Experiments}

\subsection{Multilayer Perceptron Network} 

\begin{figure}
\centering
\subfigure[] {\includegraphics[height=1.23in]{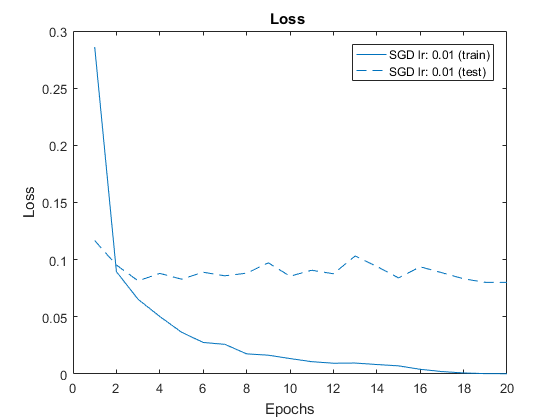}}
\subfigure[] {\includegraphics[height=1.23in]{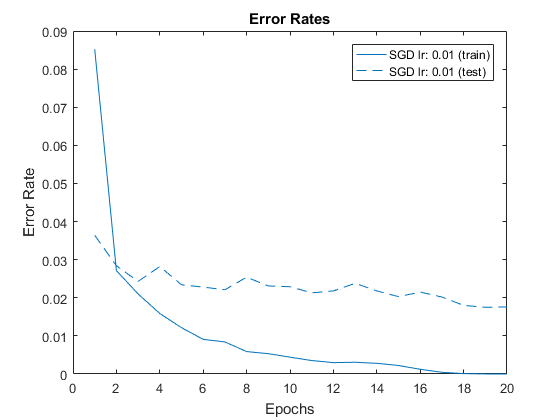}}
\caption{Loss and error rates during training and testing phases using LightNet on the MNIST dataset.}
\label{fig:MNIST-MLP}
\end{figure}

A multilayer perceptron network is constructed to test the performance of LightNet on MNIST data ~\cite{lecun1998gradient}. The network takes $28\times 28$ inputs from the MNIST image dataset and has $128$ nodes respectively in the next two layers. The $128$-dimensional features are then connected to $10$ nodes to calculate the softmax output. See Fig.~\ref{fig:MNIST-MLP} for the experiment results.

\subsection{Convolutional Neural Network} 

\begin{figure}
\centering
\subfigure[] {\includegraphics[height=1.23in]{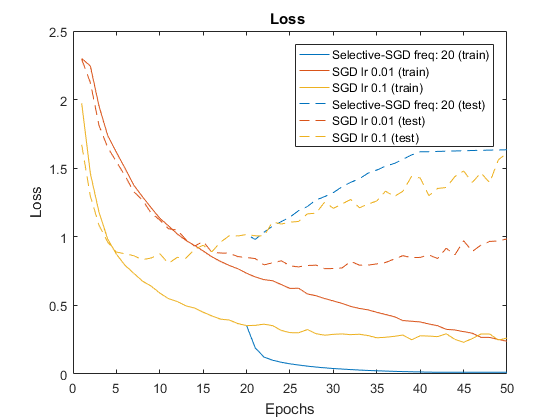}}
\subfigure[] {\includegraphics[height=1.23in]{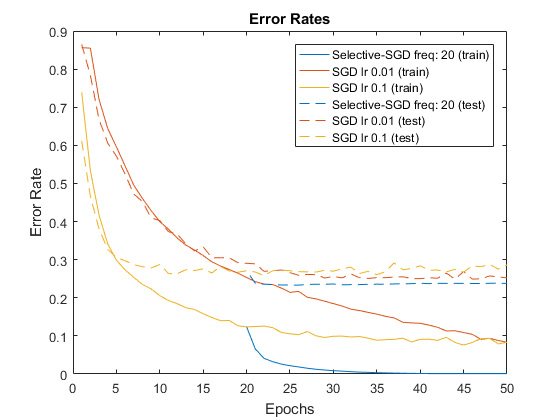}}
\caption{Loss and error rates of training and testing with LightNet on the CIFAR-10 dataset.}
\label{fig:CIFAR-10}
\end{figure}

LightNet supports using state-of-the-art convolutional network models pretrained on the ImageNet dataset. It also supports training novel network models from scratch. A convolutional network with $4$ convolution layers is constructed to test the performance of LightNet on CIFAR-10 data ~\cite{krizhevsky2009learning}. There are $32,32,64,64$ convolution kernels of size $5\times 5$ in the first three layers, the last layer has kernel size $4\times 4$. $relu$ functions are applied after each convolution layer as the non-linear mapping function. LightNet automatically selects and adjusts the learning rate and can achieve state-of-the-art accuracy with this architecture. Selective-SGD leads to better accuracy compared with standard SGD with a fixed learning rate. Most importantly, using Selective-SGD avoids manual tuning of the learning rate. See Fig.~\ref{fig:CIFAR-10} for the experiment results. The computations are carried out on a desktop computer with an Intel i5 6600K CPU and a Nvidia Titan X GPU with 12GB memory. The current version of LightNet can process $750$ images per second with this network structure on the GPU, around $5\times$ faster than using CPU.

\subsection{LSTM Network}
The Long Short Term Memory (LSTM) ~\cite{hochreiter1997long} is a popular recurrent neural network model. Because of LightNet's versatility, the LSTM network can be implemented in the LightNet package as a particular application. Notably, the core computational modules in LightNet are used to perform time domain forward process and back propagation for LSTM. 

The forward process in an LSTM model can be formulated as:
\begin{equation}
i_t= sigmoid (W_{ih} h_{t-1} + W_{ix} x_t + b_i),
\label{LSTM_i}
\end{equation}
\begin{equation}
o_t= sigmoid (W_{oh} h_{t-1} + W_{ox} x_t + b_o),
\label{LSTM_o}
\end{equation}
\begin{equation}
f_t= sigmoid (W_{fh} h_{t-1} + W_{fx} x_t + b_f),
\label{LSTM_f}
\end{equation}
\begin{equation}
g_t= tanh (W_{gh} h_{t-1} + W_{gx} x_t + b_g),
\label{LSTM_g}
\end{equation}
\begin{equation}
c_t= f_t \odot c_{t-1}+i_t \odot g_t,   
h_t= o_t \odot tanh(c_t),
\label{LSTM_c_h}
\end{equation}
\begin{equation}
z_t= f(h_t), 
z= \sum_{t=1}^{T} z_t.
\label{LSTM_loss}
\end{equation}

Where $i_t/o_t/f_t$ denotes the response of the input/output/forget gate at time $t$. $g_t$ denotes the distorted input to the memory cell at time $t$. $c_t$ denotes the content of the memory cell at time $t$. $h_t$ denotes the hidden node value. $f$ maps the hidden nodes to the network loss $z_t$ at time $t$. The full network loss is calculated by summing the loss at each individual time frame in Eq.~\ref{LSTM_loss}.

To optimize the LSTM model, back propagation through time is implemented and the most critical value to calculate in LSTM is: $\frac{\partial z}{\partial c_{s}}=\sum_{t=s}^{T}\frac{\partial z_t}{\partial c_s}$.

A critical iterative property is adopted to calculate the above value:
\begin{equation}
\frac{\partial z}{\partial c_{s-1}}=\frac{\partial z}{\partial c_{s}}\frac{\partial c_{s}}{\partial c_{s-1}}+\frac{\partial z_{s-1}}{\partial c_{s-1}}.
\label{LSTM_dzdc-2}
\end{equation}

A few other gradients can be calculated through the chain rule using the above calculation output:
\begin{equation}
\frac{\partial z_t}{\partial o_{t}}=\frac{\partial z_t}{\partial h_{t}}\frac{\partial h_t}{\partial o_{t}},
\frac{\partial z}{\partial \{ i,f,g \}_{t}}=\frac{\partial z}{\partial c_{t}}\frac{\partial c_t}{\partial \{ i,f,g \}_{t}}.
\label{LSTM_dcdifgo}
\end{equation}

The LSTM network is tested on a character language modeling task. The dataset consists of $20,000$ sentences selected from works of Shakespeare. Each sentence is broken into 67 characters (and punctuation marks), and the LSTM model is deployed to predict the next character based on the characters before. 30 hidden nodes are used in the network model and RMSProp is used for the training. After 10 epochs, the prediction accuracy of the next character is improved to $70\%$.

\subsection{Q-Network}

As an application in reinforcement learning, We created a Q-Network~\cite{mnih2015human} with the MLP network. The Q-Network is then applied to the classic Cart-Pole problem ~\cite{barto1983neuronlike}. The dynamics of the Cart-Pole system can be learned with a two-layer network in hundreds of iterations. One iteration of the update process of the Q-Network is:
\begin{multline}
Q_{new}(state_{old},act)=reward+\gamma Q_{current}(state_{new},act_{best}) \\
=reward+\gamma max_{a} Q_{current}(state_{new}, a) \\
=reward+\gamma V(state_{new}).
\label{Q_Network}
\end{multline}

The $action$ is randomly selected with probability $epsilon$, otherwise the $action$ leading to the highest score is selected. The desired network output $Q_{new}$ is calculated using the observed reward and the discounted value $\gamma V(state_{new})$ of the resulting state, predicted by the current network through Eq.~\ref{Q_Network}.

By using a least squared loss function:
\begin{multline}
z=(y-Q_{current}(state_{old},act))^2\\
=(Q_{new}(state_{old},act)-Q_{current}(state_{old},act))^2,
\label{Q_Network_loss}
\end{multline}

the Q-Network can be optimized using the gradient:
\begin{equation}
\frac{\partial z}{\partial \theta}=\frac{\partial z}{\partial Q_{current}}\frac{\partial Q_{current} }{\partial \theta}.
\label{Q-BP}
\end{equation}

Here $\theta$ denotes the parameters in the Q-Network.

\section{Conclusion}
LightNet provides an easy-to-expand ecosystem for the understanding and development of deep neural network models. Thanks to its user-friendly Matlab based environment, the whole computational process can be easily tracked and visualized. This set of the main features can provide unique convenience to the deep learning research community.


{\small
\bibliographystyle{acm}
\bibliography{references.bib}
}

\end{document}